\documentclass[10pt,twocolumn,letterpaper]{article}

\usepackage{cvpr}
\usepackage{times}
\usepackage{epsfig}
\usepackage{amssymb}

\usepackage[T1]{fontenc}    \usepackage{url}            \usepackage{amsfonts}       \usepackage{nicefrac}       \usepackage{microtype}      
\usepackage{amsmath}
\usepackage[ruled,vlined]{algorithm2e} \usepackage{graphicx}
\usepackage{lipsum}
\usepackage{booktabs}
\usepackage{array,multirow}
\usepackage{xcolor}
\usepackage[british,american]{babel}
\usepackage{blindtext}
\usepackage[inline]{enumitem}
\usepackage{balance}
\usepackage[pagebackref=true,breaklinks=true,letterpaper=true,colorlinks,bookmarks=false]{hyperref}

\cvprfinalcopy 
\def\cvprPaperID{191} 

\ifcvprfinal\fi

\newcommand{\figref}[1]{\mbox{Fig. \ref{#1}}}
\newcommand{\tblref}[1]{\mbox{Table \ref{#1}}}

\renewcommand{\eqref}[1]{\mbox{Eq. \ref{#1}}}

\newcommand{\ours}{\mbox{{Ours}~}}
 
\usepackage[font=small]{subcaption}
\newcommand{\fsz}{\fontsize{7pt}{7.0pt}\selectfont }
\let\oldparagraph\paragraph
\renewcommand{\paragraph}[1]{\vspace{-0.4cm} \oldparagraph{#1}}

\newcommand{\figvspace}{\vspace{-0.3cm}}
\newcommand{\figlblvspace}{\vspace{-0.1cm}}
\newcommand{\eqtopvspace}{\vspace{-0.1cm}}
\newcommand{\eqbottomvspace}{\vspace{-0.0cm}}
\newcommand{\tabvspace}{\vspace{-0.3cm}}
\newcommand{\tablblvspace}{\vspace{-0.1cm}}
\newcommand{\reffontsize}{\small} 

\begin{document}

\title{Automatic Discovery, Association Estimation and Learning of \\Semantic Attributes for a Thousand Categories}

\author{Ziad Al-Halah \hspace{2cm} Rainer Stiefelhagen \\
Karlsruhe Institute of Technology, 76131 Karlsruhe, Germany\\
{\tt\small \{ziad.al-halah, rainer.stiefelhagen\}@kit.edu}
}

\maketitle

\begin{abstract}
\makeatletter{}Attribute-based recognition models, due to their impressive performance and their ability to generalize well on novel categories,  have been widely adopted for many computer vision applications. 
However, usually both the attribute vocabulary and the class-attribute associations have to be provided manually by domain experts or large number of annotators. 
This is very costly and not necessarily optimal regarding recognition performance, and most importantly, it limits the applicability of attribute-based models to large scale data sets. 
To tackle this problem, we propose an end-to-end unsupervised attribute learning approach.
We utilize online text corpora to automatically discover a salient and discriminative vocabulary that correlates well with the human concept of semantic attributes. 
Moreover, we propose a deep convolutional model to optimize class-attribute associations with a linguistic prior that accounts for noise and missing data in text. 
In a thorough evaluation on ImageNet, we demonstrate that our model is able to efficiently discover and learn semantic attributes at a large scale.  Furthermore, we demonstrate that our model outperforms the state-of-the-art in zero-shot learning on three data sets: ImageNet, Animals with Attributes and aPascal/aYahoo.
Finally, we enable attribute-based learning on ImageNet and will share the attributes and associations for future research.
 
\end{abstract}

\section{Introduction}
\makeatletter{}
\begin{figure}[t]
\centering
\includegraphics[width=0.99\linewidth]{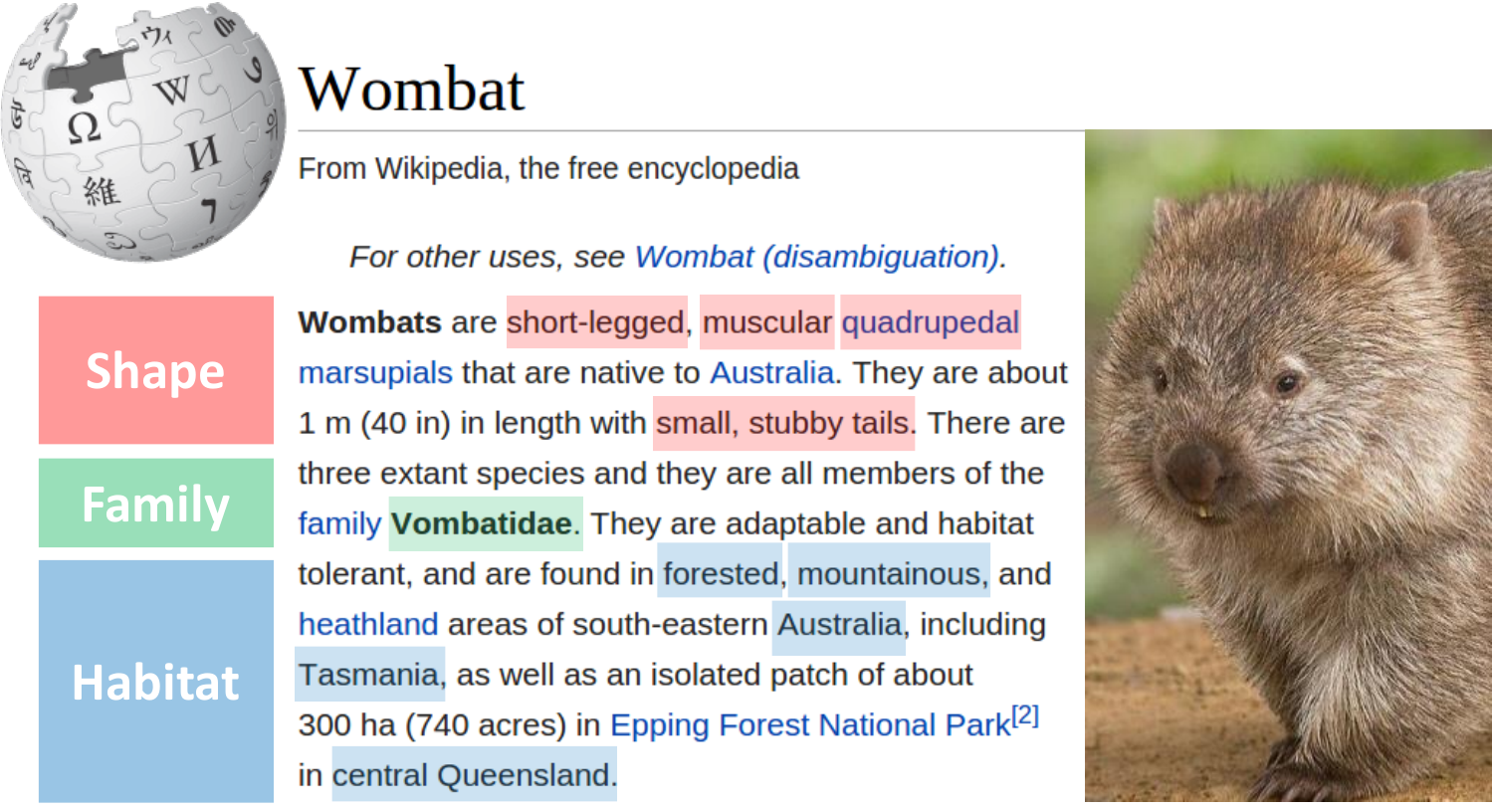}
\figlblvspace
\caption{An encyclopedia article describing an object category. 
Many discriminative attributes regarding shape, family and habitat of the object can be identified already in the first few lines of the article.
We propose a model that utilizes such a knowledge source to automatically discover and learn visual semantic attributes at a large scale.}
\label{fig:intro}
\figvspace
\end{figure}
 
\makeatletter{}
Semantic attributes, being both machine detectable and human understandable, lend themselves to various applications in vision and language domain \cite{Rohrbach2010}.
By creating an intermediate layer of semantics that cross the boundaries of object categories, they also show impressive performance in transfer learning \cite{Lampert2009} and domain adaptation \cite{Chen2015}. 
However, attribute annotations for object categories are usually obtained manually by tens of annotators \cite{Farhadi2009,Lampert2013} or domain experts \cite{Wah2011}.
Moreover, the attribute vocabulary itself requires careful engineering.
It should be shared across the categories but at the same time be discriminative, interpretable and visually detectable.

This is clearly a major obstacle for attribute-based approaches to scale to large number of classes.
The cost associated in providing such annotations is prohibitive which limits the available attribute data sets either in the number of classes, attributes or images.
Additionally, this non-trivial and expensive work is needed again when moving across data sets or expanding the current set with new categories.

We aim in this work to circumvent this need for human intervention.
Our goal is to automatically mine attributes vocabulary and find their associations to objects in a large scale setting.
We achieve this by utilizing the large text corpora available in the web.
Online encyclopedias represent a rich source of information which encode the collective human knowledge over various concepts and categories.
It is an active and comprehensive knowledge source that keeps growing at impressive rates \cite{WikiSize}. 
\figref{fig:intro} shows a snippet of an article describing the animal category \emph{Wombat}.
One can easily observe that numerous distinctive attributes for this category about its shape, taxonomy and habitat already appear in the introduction of the article.

Recently, This valuable and massive source of knowledge attracted a lot of interest in the vision community.
One can identify two main perspectives in that direction. 
The first, learns word embeddings of the object categories from text using powerful models from natural language processing \cite{Mikolov2013,Pennington2014}.
It then adapts a deep neural model for embedding prediction and zero-shot learning based on visual data \cite{Frome2013,Norouzi2014}.
Differently, the second perspective follows a domain adaptation approach between language and vision.
It directly predicts classifier weights for unseen classes based on an embedding of the category textual description \cite{Elhoseiny2013,Ba2015,Qiao2016}.
While we tap to a similar knowledge source to bridge the gap between language and vision, in contrast to these approaches, our objective is to automatically discover an explicit set of semantic attributes that is compact, discriminative and best describes the categories in our data. 

\paragraph{Contributions} 
The main contributions of our work are as follows:
\begin{enumerate*}[label={\alph*)}]
\item We propose a novel attribute mining approach from natural textual description that not only accounts for discrimination but also mines a diverse and salient vocabulary which correlates well with the human concept of semantic attributes.
\item We propose a novel approach to associate these mined attributes with classes using a deep convolutional model that leverages visual data to account for the noisy and missing information in the text corpora.
\item We experimentally demonstrate that our deep attribute model is able to learn and predict attributes with high accuracy on ImageNet, as well as it generalizes well across data sets and outperforms state of the art in zero-shot learning on three benchmarks.
\item Finally, as a result of our work, we have collected textual descriptions for more than a thousand categories; furthermore, we have automatically generated attribute annotations for ImageNet and deep attribute models that will be made publicly available\footnote{ \url{http://cvhci.anthropomatik.kit.edu/~zalhalah/}}. 
We believe, this data might be of great interest for the vision and language research community.
\end{enumerate*}

\section{Related work}
\makeatletter{}While attribute-based visual recognition gained a continuous rise in popularity in computer vision, collecting attribute annotations proved to be very expensive.
Consequently, this clearly limits the scalability of attribute-based approaches to large number of categories.
Most available attribute data sets \cite{Farhadi2009,Lampert2009,Wah2011} are limited in terms of the number of attributes, categories or images.
In a recent effort to collect a larger attribute data set, \cite{Patterson2016} proposed a cost effective labeling approach where they collected annotations of 196 attributes for 29 categories and 84 thousand images with annotation cost of more than \$30,000.
Differently, in this work, we circumvent the need of user supervision to define and annotate attributes.
We propose an unsupervised end-to-end approach to automatically mine and learn semantic attributes for thousands of categories.

\paragraph{Attribute discovery}
There were few attempts in the literature to automatically obtain an attribute vocabulary.
\cite{Rohrbach2010,Rohrbach2011} mine attributes by crawling the WordNet \cite{Miller1995} ontology.
Specifically, they track the ``has-part'' relations in WordNet to extract ``part'' attributes. On the other hand, \cite{Ferrari2008,Chen2013,Divvala2014} use the top ranked images returned by web search engines queried with a certain vocabulary to estimate the ``visualness'' of words. 
\cite{Berg2010} samples pairs of (image, description) from the Internet to automatically find a set of visual attributes. 
Similarly, \cite{Vittayakorn2016} uses both image-based textual descriptions as well as a set of image tags provided by users in social media to identify the attribute vocabulary.
Different from previous work, our approach does not require images aligned with textual descriptions or tags.
Furthermore, we do not rely on a predefined ontology such as WordNet or target only a specific type of attributes like ``parts''.
Instead, we use textual description at the category level in form of encyclopedia entries to extract a salient and diverse set of attributes.

\paragraph{Class-attribute association prediction}
In a different direction, other approaches focused on predicting the class-attribute associations automatically. 
\cite{Rohrbach2010,Mensink2014} estimate the associations strength from web-based co-occurrence statistics.
However, web-based hit counts estimations are noisy since it does not take into consideration the context or the specific relation sought between the category and the attribute. 
\cite{Al-Halah2015} uses WordNet hierarchy to transfer the attribute associations of an unseen class from its parent in the ontology.
Recently, \cite{Al-Halah2016} proposes to predict attribute associations using semantic relations in a tensor factorization approach. 
However, both \cite{Al-Halah2015} and \cite{Al-Halah2016} assume the availability of training associations.
Here, we propose a deep model to estimate the class-attribute associations from scratch.
Our model takes advantage of an initial linguistic prior over the associations from textual description and improves the estimations in a joint optimization framework of object and attributes predictions.

\paragraph{Unsupervised zero-shot learning (ZSL)}
Semantic attributes -- with their ability to be shared across categories -- have shown impressive performance in tasks like ZSL.
However, due to their limited scalability, there is an increasing interest in conducting ZSL by tapping to an alternative knowledge source, for example by exploiting lexical hierarchies to transfer visual models between the categories \cite{Rohrbach2011} or learning a hierarchical embedding \cite{Akata2015}.
A different direction leverage powerful word embeddings~\cite{Huang2012,Mikolov2013} to establish the semantic link between seen and unseen categories \cite{Frome2013,Norouzi2014}.
More closely to our work are the ones from \cite{Elhoseiny2013,Ba2015} and \cite{Qiao2016}.
These approaches use article embeddings to directly predict the classifier weights of the novel category in a domain adaption framework.
However, most of the unsupervised ZSL approaches do not result in good discriminative classifiers when compared to their attribute-based counterpart \cite{Al-Halah2016}.
We show in the evaluation, that our unsupervised deep attribute model can predict novel categories with high accuracy and it outperforms state-of-the-art in unsupervised ZSL with a significant margin.

\section{Discovering and Learning Attributes}
\makeatletter{}We propose an end-to-end approach for large scale attribute-based visual recognition.
Starting with a set of articles describing the object categories, our approach consists of three main steps: 
1) We automatically analyze the articles in order to extract an attribute vocabulary with the most salient and discriminative words to describe these categories.
Then, 2) we optimize the class-attribute associations using visual data by a novel deep convolutional model with a linguistic prior and joint optimization of class and attribute predictions.
Finally, 3) we train a deep neural model for large scale attribute classification. 

\subsection{Semantic attribute discovery}\label{sec:word_sel}
\makeatletter{}
\begin{figure}[t]
\centering
\includegraphics[width=0.98\linewidth]{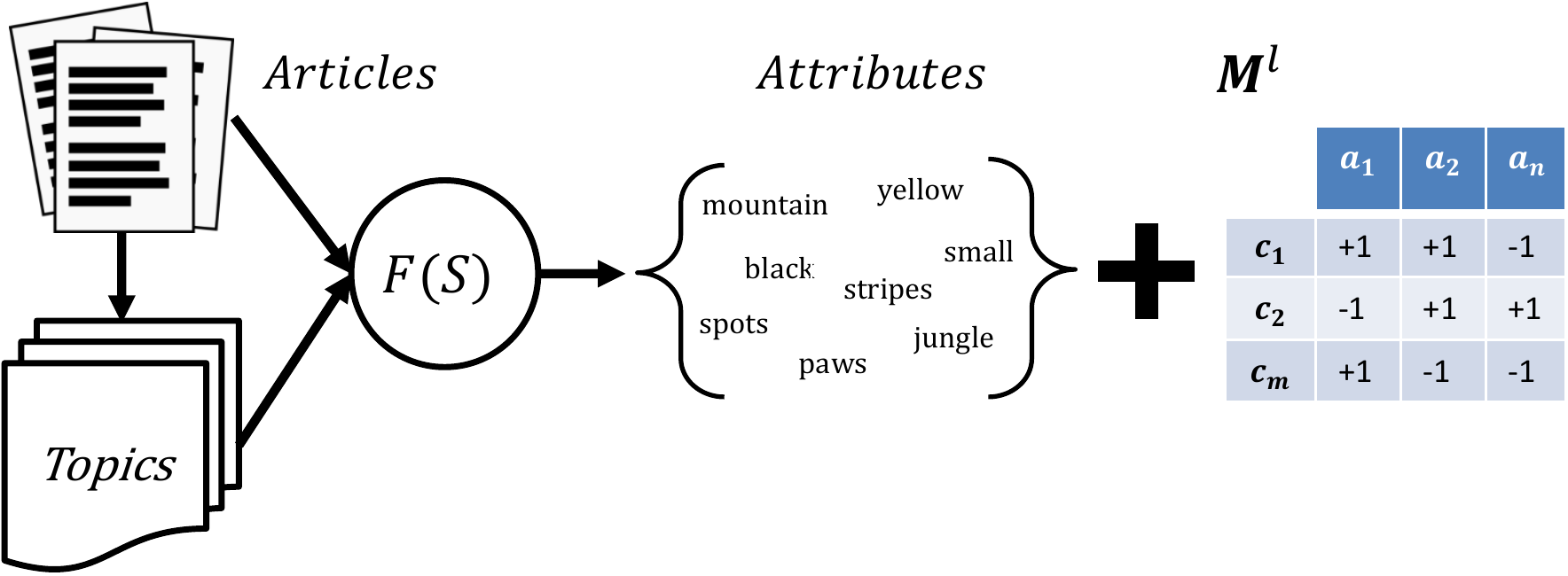}
\figlblvspace
\caption{Discovering an attribute vocabulary from textual descriptions. Our model leverages text from articles and their underlying latent topics to select a compact, discriminative, diverse and salient set of semantic attributes.}
\label{fig:model_1}
\figvspace
\end{figure}
 
\makeatletter{}Let $D=\{d_j\}_{j=1}^J$ be a set of text documents describing all object categories $C=\{c_m\}_{m=1}^{M}$ in the dataset.
For notation simplicity, we assume $|D|=|C|$, \ie there is one document for each category. 
Let $W=\{w_i\}_{i=1}^{I}$ be the dictionary of words learned from $D$.
Then, our goal is to select a subset vocabulary $A \subseteq W$ that best describes $C$:
\begin{equation}\label{eq:max_f}
A = \arg \max_{S \subseteq W} \mathcal{F}(S) \, \mathrm{where}\, |S|\leq b,
\end{equation}
where $\mathcal{F}$ is a set function that captures the desired properties of the subset $S$, and $b$ is the size of the vocabulary.

Ideally, words in $A$ should: 1) discriminate well between the object categories; 2) describe diverse aspects of the categories rather than focusing only on one or few properties (e.g. only colors or parts); and 3) represent salient semantic concepts understandable by humans.
Next, we describe how we capture these different criteria of $S$ in our objective function (\figref{fig:model_1}). 

\paragraph{Discrimination}
Let $V=\{\mathbf{v}_j=f_v(d_j): \mathbf{v}_j\in\mathbb{R}^{|W|}\}_{j=1}^J$ be a text-based embedding (\eg $f_v(\cdot)$ is based on tf$\cdot$idf) learned over the document set $D$ such that $v_j^i$ captures the word $w_i$ importance in document $d_j$.
We construct an undirected fully connected graph $\mathcal{G}(N,E)$.
Each node $n_i \in N$ represents a category $c_i$.
Each edge $e_{ij} (i\neq j)$ has a weight $g_{ij}(S)= \sum_{w_k \in S} |v_i^k - v_j^k|$ that captures how well words in $S$ discriminate one class from the others.
Additionally, each node has a self loop $e_{ii}$ with a weight $g_{ii}(S)= \sum_{j\neq i}\sum_{w_k \notin S} |v_i^k - v_j^k|$.
To capture the discriminative power of a set $S$, we employ the entropy rate of a random walk $\mathcal{X}$ on graph $\mathcal{G}$ as defined by \cite{Liu2014,Zheng2014}.

In summary, let $g_{i}(S)=\sum_j g_{ij}(S)$ be the sum of incident weights of node $n_i$ and the total sum of weights in the graph is $g_T=\sum g_i$.
The transition probability among the nodes is set to: \begin{equation}\label{eq:transition}
\eqtopvspace
p_{ij}(S) = \left\{
  \begin{array}{l l}
    \frac{g_{ij}(S)}{g_{i}(S)} & \, \mathrm{if} \, i\neq j\\
    1 - \frac{\sum_j g_{ij}(S)}{g_{i}(S)} & \, \mathrm{if} \, i = j\\
  \end{array}\right.
\eqbottomvspace
\end{equation}
Note that $p_{ij}$ is a set function and the transition probabilities will change when the selected set $S$ changes.
The incident weights $g_i$ for each node in the graph are kept constant because of the self loops weight $g_{ii}$, and the stationary distribution for the random walk is defined as $\mu=(\mu_1,\mu_2,\dots,\mu_{|N|})$, where $\mu_i=\frac{g_{i}}{g_T}$.
Then the entropy rate of a random walk on $\mathcal{G}$ is:
\begin{equation}\label{eq:discrimiation}
\mathcal{F}_{dis}(S) = -\sum_i \mu_i \sum_j p_{ij}(S)log(p_{ij}(S))
\end{equation}
The maximization of $\mathcal{F}_{dis}$ demands the maximization of $p_{ij}$ \ie the discrimination among all pairs of classes.

\paragraph{Diversity}
Another desired property of a good set of attributes is that it describes various aspects of the categories. 
That is, we want to encourage diversity among the selected words to reduce the bias towards a specific set of classes and to mine a vocabulary that describes all categories equally well.
In order to promote diversity, we first uncover the latent semantic structure among the categories.
We leverage here the unsupervised probabilistic topic models (\eg LDA \cite{Blei2003}) to discover underlying themes in the documents.

Let $T=\{T_k\}_{k=1}^{K}$ be a set of topics learned from documents $D$ and dictionary $W$.
We define the diversity objective criteria as:
\begin{equation}\label{eq:diversity}
\eqtopvspace
\begin{split}
\mathcal{F}_{div}(S) &= \sum_{T_k} \sqrt{\sum_{w_i \in S} s(w_i,T_k)} \quad \mathrm{where}\\
 s(w_i,T_k) &= \left\{
  \begin{array}{l l}
    p(w_i|T_k)  & \, \mathrm{if} \, T_k = \arg\max\limits_{T_j} p(T_j|w_i)\\
    0 			& \, \mathrm{otherwise}\\
  \end{array}\right.
\end{split}
\eqbottomvspace  
\end{equation}

$\mathcal{F}_{div}$ encourages topic diversity in $S$ since adding words that belong to a previously chosen topic will have diminishing gain because of the square root function. 
It also accounts for word importance for the topic since adding a word with higher $p(w_i|T_k)$ results in a higher gain.
Moreover, $\mathcal{F}_{div}$ by encouraging diversity also discourages redundancy. 
A word and its synonyms are more likely to belong to the same topic, hence they are less likely to be chosen together. 
That is, $\mathcal{F}_{div}$ favors a diverse, less redundant and representative set of words.

\paragraph{Saliency}
An important aspect of semantic attributes is that they represent \emph{salient} words with relatively clear semantic concepts, \eg ``leg'', ``yellow'' and ``transparent''.
Whereas words like ``become'', ``allow'' and ``various'' belong to the background language structure, hence they are usually ambiguous and carry less or no semantics by themselves.
Capturing word saliency directly is hard due to word polysemy and since word importance depends on the context.
Therefore, we propose to capture this property using the learned underlying topic structure among the documents as a proxy.

One can estimate the significance of a topic by comparing its distribution over the words $p(w|topic)$ and documents $p(d|topic)$ to \textit{junk} topics prototypes \cite{AlSumait2009}.
A junk topic is one that has uniform distribution over words (\ie it doesn't capture any specific theme) or over documents (\ie it captures the common theme of all documents).
By measuring the distance (\eg KL divergence) of the discovered topics to these \textit{junk} prototypes, we can obtain a ranking of the topics regarding their significance.

\makeatletter{}
\begin{table}
\renewcommand{\arraystretch}{0.5}
\centering
\scalebox{0.9}{{
\begin{tabular}{c p{6cm}}
\toprule
Rank 	& Top Words in Topic\\ 
\midrule 
  1 	& \fsz instrument play music sound pitch note musical reed player violin make tone range octave bass family key band fiddle hole \\ 
  2 	& \fsz spaniel english welsh cocker springer show cardigan field pembroke work dock type small sussex average come line variety would century \\ 
  3 	& \fsz missile target system wing guide flight use force parachute engine know projectile rocket air lift guidance kinetic anti weapon shuttle \\ 
 $\vdots$ &    \\ 
198 	& \fsz call include allow many time upper consist long much several little last low reach second slow half make follow suitable \\ 
199 	& \fsz use make allow would prevent work take give open cause come reduce keep provide way protect help less leave property \\ 
200 	& \fsz use century become modern early world work time begin develop could history new war late development introduce part include today \\ 

\bottomrule
\end{tabular}}}
\tablblvspace
\caption{Ranking of discovered topics according to their significance, \ie how different they are from \emph{junk} topic prototypes.
While the top ranked topics capture salient concepts like \emph{music} and \emph{dogs}, the low ranked ones are obscure and have no particular theme.}
\label{tab:topic_rank}
\tabvspace
\end{table} 
\tblref{tab:topic_rank} shows the highest and lowest ranked topics over a set of documents using topic significance analysis model from \cite{AlSumait2009}.
One can see that the top ranked topics revolve around specific themes like ``music'', ``dogs'' and ``military'', while the lowest ranked topics have no theme in particular and are related to the background structure of the language or the documents domain.

Let $\mathrm{insig}(T)$ be the set of $\rho=10\%$ lowest ranked topics. 
We define a saliency cost function as: \begin{equation}\label{eq:saliency}
\mathcal{C}(S) = \sum_{w_i \in S} (1 + \gamma \sum_{T_k \in \mathrm{insig}(T)} p(T_k|w_i)),
\end{equation}
where $\gamma$ controls the contribution of the insignificance score of a word to the cost function. 
$\mathcal{C}(\cdot)$ favors \textit{salient} words which will have a cost close to 1 while it punishes \textit{junk} words which have a higher probability to appear in \textit{junk} topics. 

\paragraph{Submodular optimization}
We formulate the vocabulary selection problem in a submodular knapsack framework \cite{Atamturk2009}.
A set function $\mathcal{F}$ is submodular if it satisfies the decreasing marginal gain condition \cite{Fujishige2005} \ie: $\mathcal{F}(A\cup\{s\})-\mathcal{F}(A) \geq \mathcal{F}(B \cup \{s\}) - \mathcal{F}(B) ~\mathrm{for}~ A\subseteq B$.
In other words, the benefit of adding a new element $s$ to the set is higher if it happens earlier.
All the previous functions $\mathcal{F}_{dis}$, $\mathcal{F}_{div}$ and $\mathcal{C}$ satisfy the marginal gain condition and are submodular\footnote{more details in supplementary}.
We formulate our main objective function as:
\newcommand{\sm}{\hspace{8pt}}
\begin{equation}\label{eq:objective}
\begin{array}{l l}
 \max\limits_{S \subseteq W} \mathcal{F}(S) &= \mathcal{F}_{dis}(S) + \lambda \mathcal{F}_{div}(S) \\
 &\mathrm{subject~to} \sm \mathcal{C}(S) \leq b
\end{array}
\end{equation}

where $b$ is the budget and $\lambda$ a hyper-parameters controlling the contribution of $\mathcal{F}_{div}$.
 
$\mathcal{F}(\cdot)$ is submodular since it is a linear combination of submodular functions \cite{Fujishige2005}.
Submodular functions can be optimized robustly with a guaranteed solution to be near optimal \cite{Krause2008}.
We adopt a lazy greedy algorithm \cite{Leskovec2007}.
We start with an empty set $S=\{\}$, then we incremently add elements to $S$ with maximum gain according to $\mathcal{F}$ using lazy evaluations.

\subsection{Association optimization with a linguistic prior}\label{sec:assoc_opt}
\makeatletter{}
\begin{figure}[t]
\centering
\begin{subfigure}[t]{0.47\linewidth}
    \centering
    \includegraphics[width=\linewidth]{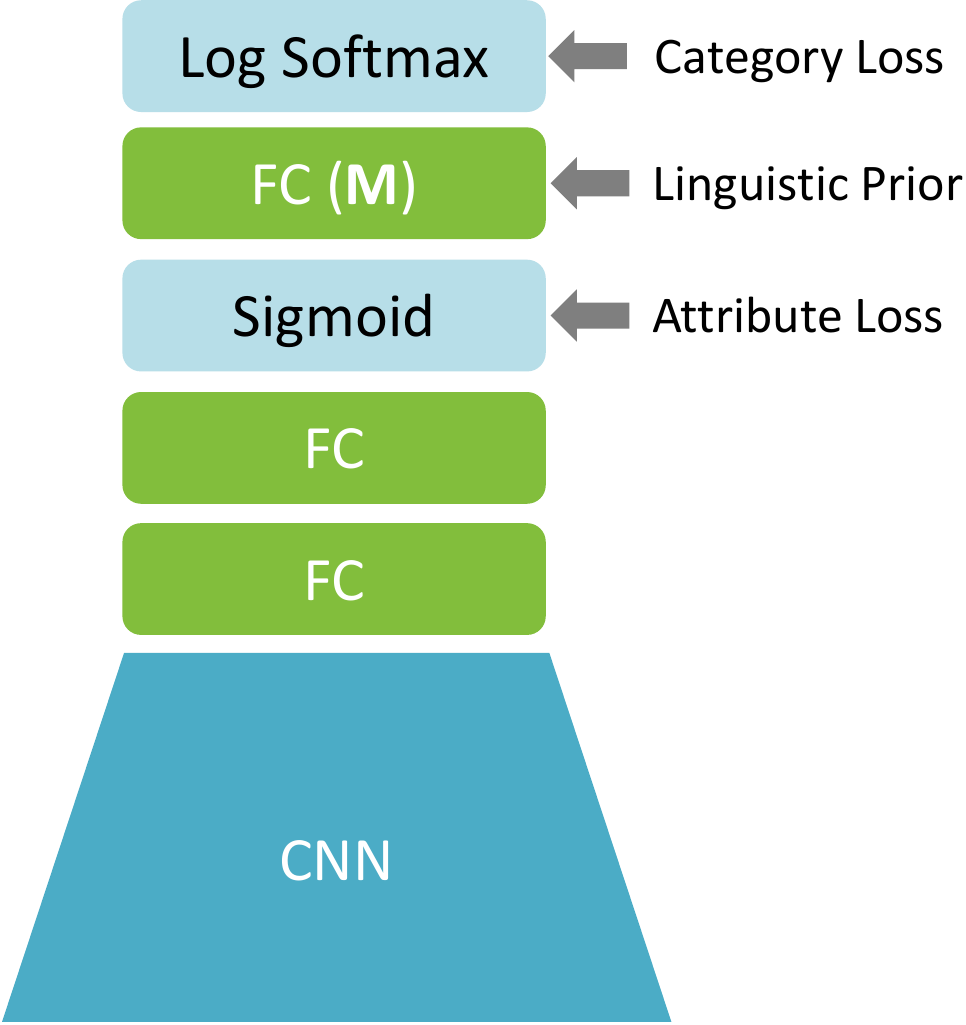}
    \caption{}\label{fig:model_2}
\end{subfigure}\quad
\begin{subfigure}[t]{0.47\linewidth}
    \centering
    \includegraphics[width=\linewidth]{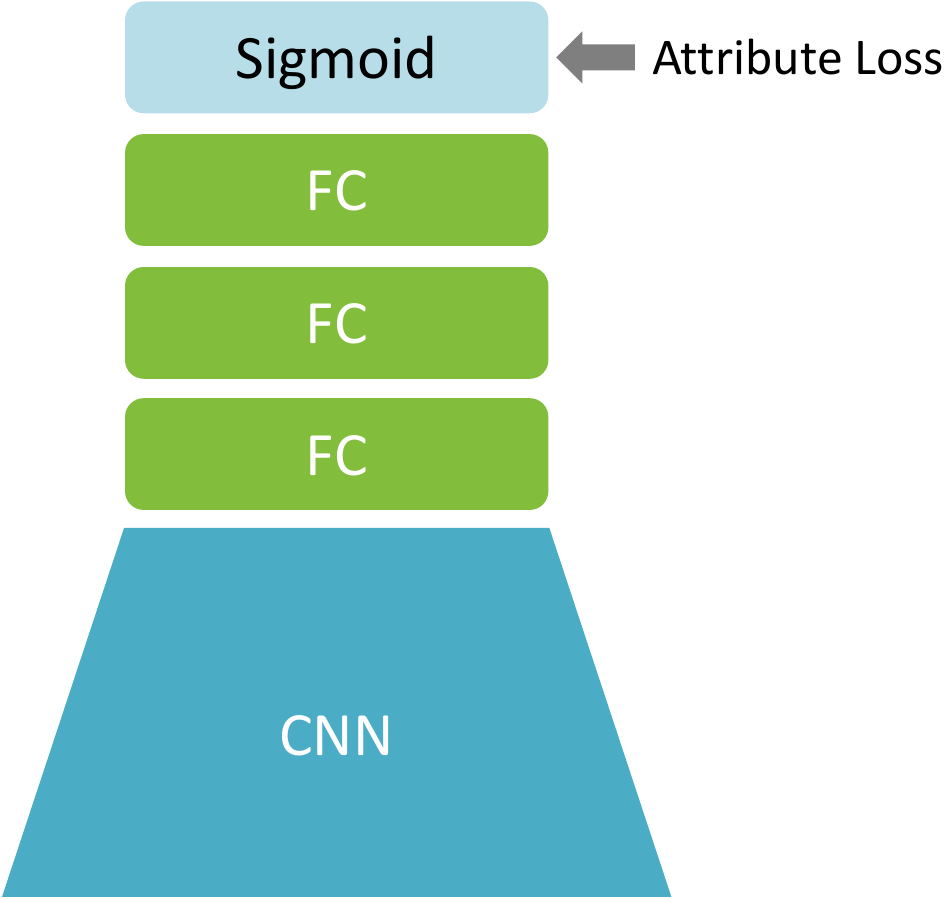}
    \caption{}\label{fig:model_3}
\end{subfigure}
\figlblvspace
\caption{(a) The joint optimization of class-attribute associations using a linguistic prior and (b) the deep attribute model architecture.}
\label{fig:model_2_3}
\figvspace
\end{figure}
 
\makeatletter{}In the previous step, we have selected the best attribute vocabulary $A$ that describes the different categories $c_i \in C$ in our data set.
Having this set of words, we get an initial estimate of the class-attribute association matrix $\mathbf{M}^l=[m_{ij}]$ (\figref{fig:model_1}) using the text-based embedding $\mathbf{V}$ learned over $D$. 
\begin{equation}
\eqtopvspace
m_{ij} = \left\{
  \begin{array}{l l}
    +1 & \, \mathrm{if} \, v^i_j > 0\\
    -1 & \, otherwise\\
  \end{array} \right.
\eqbottomvspace
\end{equation}
However, this association matrix may contain some noise since $\mathbf{V}$ does not capture context, and not all relations for a certain category are necessarily represented in the respective text documents. 
Usually, simple and obvious attributes of a class are omitted from text if they are not interesting enough to mention from the perspective of the author.
For example, while most animals have attributes like ``head'', ``leg'' or ``skin'' these are not always mentioned in text when describing the animal unless there is something special about it.
Moreover, $\mathbf{V}$ is a bag of words representation, \ie it does not capture the context of the attributes in text.
This results in a negative relation like ``a tiger does not live in ocean'' being captured as a positive association between ``tiger'' and ``ocean'' since $\mathbf{V}$ relies only on the presence of the word in the description.

We propose to improve the initial associations obtained from language by grounding it to visual data using a deep convolutional network model.
The network is trained to predict both attributes and categories while at the same time constraining the weights of the last layer to the initially estimated associations $\mathbf{M}^l$ (see \figref{fig:model_2}). 
Note that this architecture resembles the direct attribute prediction model DAP \cite{Lampert2009} where the object class is estimated based on the predicted attributes.
We define the training loss function as: 
\begin{equation}
\mathcal{L}(x) = \mathcal{L}_c (x) + \beta_1 \mathcal{L}_a (x) + \beta_2 \|\mathbf{M} - \mathbf{M}^l\|_1 \end{equation}
where $\mathcal{L}_c$ and $\mathcal{L}_a$ are the cross entropy loss of predicting the object category and the binary attributes of sample $x$, respectively. 
$\|\mathbf{M} - \mathbf{M}^l\|_1$ is an entry-wise $\mathrm{L}_1$ regularization term over the weights of the last fully connected layer $\mathbf{M}$ based on the initial association matrix $\mathbf{M}^l$.

Note that, by using the linguistic prior we force the network to preserve the semantic link between linguistic and visual data.
This prevents the network from finding arbitrarily data-driven associations that can not be estimated anymore from textual description.
At the same time, by controlling $\beta_2$ we allow for small modification to the associations when there is a strong visual signal supporting change to account for noise and missing information in $\mathbf{M}^l$. 

We adopt an AlexNet-like architecture \cite{Krizhevsky2012} for the joint deep model.
That is, we have 5 convolutional layers followed by two fully connected layers and a Sigmoid activation function for attribute prediction, then another fully connected layer with softmax activation for category classification.
At the end of the joint optimization, we get the new binary association matrix of classes and attributes $\mathbf{M}^*$ by thresholding the weights of the last layer $\mathbf{M}$.
The optimized associations $\mathbf{M}^*$ redefine the positive and negative label assignments for each attribute which were intially based on $\mathbf{M}^l$.

\subsection{Deep attribute model}\label{sec:attr_model}
\makeatletter{}Finally, given the optimized associations $\mathbf{M}^*$ from the previous step, we train a deep model for attribute prediction (\figref{fig:model_3}).
The network has a similar architecture as the one we used for the joint optimization.
However, we remove the last layer for the category prediction and add a new fully connected layer before the attribute prediction layer.
That is, the network is made of 5 convolutional layers followed by three fully connected layers.
The last attribute prediction layer is followed by a Sigmoid activation function. We use the cross entropy loss to train the network for binary attribute prediction.

\paragraph{Predicting objects}
Given an image $x$, we estimate the corresponding object category using the direct attribute prediction model (DAP) \cite{Lampert2009}.
We adopt a summation formulation rather than the probabilistic one \cite{Lampert2009} since it's more efficient \cite{Rohrbach2011,Al-Halah2015}, especially in our large scale case.
That is, for a class $c_m$, the estimated prediction score of $c_m$ to appear in image $x$ as:
\begin{equation}\label{eq:dap}
s(c_m|x) = \frac{\sum_i s(a_i|x)^{a_i^{c_m}}}{\sum_i a_i^{c_m}}
\end{equation}
where $s(a_i|x)$ is the prediction score of attribute $a_i$ in image $x$, $a_i^{c_m}$ are the attributes of class $c_m$, and the classification scores are normalized to have a zero mean and unit standard deviation.
We use the same formulation for classifying unseen categories in zero-shot learning.
However, in this case the associations of the novel class are estimated directly from the textual description.

\section{Evaluation}
\makeatletter{}In this section, we provide a thorough evaluation of our model in selecting a set of good attributes, association optimization and predicting semantic attributes.
Furthermore, we evaluate our deep attribute model in zero-shot learning and its generalization properties across data sets.

\paragraph{Data setup}
Through our experiments, we use the ILSVRC2012 dataset from ImageNet \cite{ILSVRC15}. 
It contains $1000$ categories and more than $1.2$ million images.
We collect articles for each synset in the data set by querying the Wikipedia API with the different terms in each synset.
This results in 1100 articles with around 80500 unique words.
All document are preprocessed to remove non alphabetic characters, and words are lower cased and stemmed.
To avoid bias toward lengthy articles for some categories, we truncate the articles length to a maximum of 500 words.
We extract a tf$\cdot$idf (term frequency$\cdot$inverse document frequency) embedding for each document in the set.
The tf$\cdot$idf measures the importance of a word in a document by accounting for how often this word appears in the document and how frequent it appears in all other documents.
We use the normalized \emph{tf} and logarithmic \emph{idf} scores \cite{Salton1988}.
For each synset, we average the embedding over all its documents to get its final representation.

\paragraph{Implementation details}
For the attribute discovery, we learn a set of 200 topics using the Latent Dirichlet Allocation model \cite{Blei2003}.
We empirically set $\lambda=0.001$, $\gamma=20$ and the maximum number of attributes to discover $b=1200$.
We set the hyperparameters $\beta_1$ and $\beta_2$ for the joint deep model such that the initial losses from the three terms are of  similar magnitudes.
For the final deep attribute model, we initialize the weights of the convolutional layer from the previous network trained for the joint optimization.
All networks are trained using Adam \cite{Kingma2015} for stochastic optimization with an initial learning rate of 0.001 and a weight decay of 5e-4.

\subsection{Selecting the attribute vocabulary}
We evaluate the quality of the selected attribute vocabulary from two perspectives: 1) the performance of the attribute embedding in capturing object similarity and 2) the vocabulary saliency.

\makeatletter{}
\begin{figure}[t]
\centering
\includegraphics[width=0.7\linewidth]{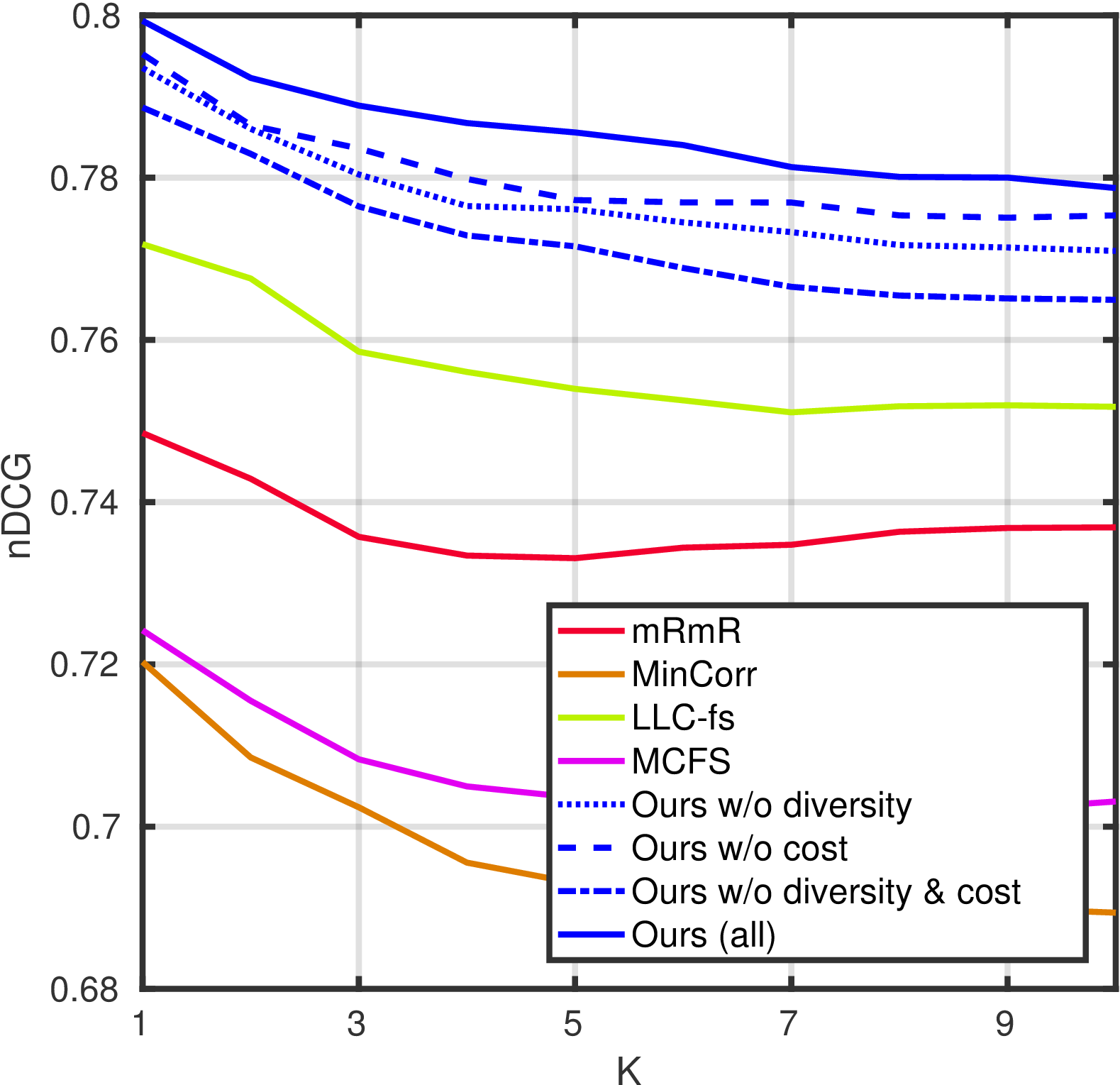}
\figlblvspace
\caption{The ranking performance of the attribute embedding from our approach against the baselines.}
\label{fig:ndcg}
\figvspace
\end{figure}
 
\paragraph{Attribute-based class embedding}
A good attribute representation of categories should capture the similarity among the classes. 
That is, categories that are visually similar should share most of their attributes and have similar embeddings.
To capture the quality of the attribute embedding, we rank the classes based on their similarity in the attribute embedding space.
We use the normalized discounted cumulative gain (nDCG) \cite{Siddiquie2011} to compare among the different methods:
\begin{equation}
\mathrm{nDCG}= \frac{DCG_k}{IDCG_k} ~~ \mathrm{where} ~~ \mathrm{DCG_k} = \sum_{i=1}^k\frac{2^{rel_i}-1}{log_2(i+1)} 
\end{equation}
Such that $rel_i$ is the relevance of the $i^{th}$ ranked sample, and the ideal rank score $\mathrm{IDCG_k}$ is that for the rank of the classes for each category based on their distances in the ImageNet hierarchy. 

As baselines, we consider several common feature selection methods: 1) max-Relevance and min-Redundancy (mRmR) \cite{Peng2005}; 2) Multi-Cluster Feature Selection (MCFS) \cite{Cai2010}; 3) Local Learning-based Clustering method (LLC-fs) \cite{Zeng2011}; 4) Minimum Correlation (MinCorr) which selects words that have the least correlation with the rest of the vocabulary. 

\figref{fig:ndcg} shows the ranking quality of all the baselines and our approach up to position K=10 in the ranking list. 
Our approach outperforms all baselines and produces an embedding that captures the within category similarities.
We also consider different variants of our approach by removing some of the optimization terms from \eqref{eq:objective}. 
Each of the terms used in our submodular optimization contributes positively to the quality of the attribute embedding. 

\makeatletter{}
\begin{table}[t]
\centering
\scalebox{0.9}{
\begin{tabular}{l c c c}
\toprule
Model 				&  Relevance (\%) $\uparrow$	&	Junk (\%) $\downarrow$ &	Saliency (\%) $\uparrow$	\\ \midrule
                mRmR &      20.8 &   53.0 &   33.9\\
             MinCorr &      14.4 &   20.6 &   46.9\\
              LLC-fs &      29.1 &   42.9 &   43.1\\
                MCFS &      18.6 &   13.6 &   52.5\\
          \midrule
                Ours &	 \textbf{44.5} &	  \textbf{2.6} &	 \textbf{71.0}\\ \bottomrule

\end{tabular}
}
\tablblvspace
\caption{Saliency scores of the selected vocabularies.}
\label{tab:words_perf}
\tabvspace
\end{table}
 
\paragraph{Vocabulary saliency}
Here, we explore how the selected vocabulary correlates with human understanding of salient semantic attributes.
To that end, we pick $100$ synsets that are uniformly distributed in the ImageNet hierarchy.
For each category, we select $50$ random words from the dictionary with positive tf$\cdot$idf scores for that class.
We asked $5$ annotators to classify the association between each class and its $50$ words into $4$ categories: 1) \emph{positive}: such as ``The horse has a tail''; 2) \emph{negative}: like ``The dolphin does not walk''; 3) \emph{unknown}: when the annotator does not have the knowledge to decide the type; 4) \emph{junk}: when the word itself does not carry a clear concept to define an association.
The majority of the annotators agree on $84\%$ of the labels.
The labels are distributed as ($25.1\%$ \emph{positive}, $47.8\%$ \emph{negative}, $1.9\%$ \emph{unknown} and $25.2\%$ \emph{junk}).
Out of the 4 categories, we are interested in the \emph{positive} and \emph{junk} categories since they describe the semantic saliency of the words.
The \emph{negative} and \emph{unknown} types do not deliver much information about the semantics since a word having a negative association might have a positive one with other classes while the \emph{unknown} reflects the lack of knowledge of the annotator.
We obtain the probability of a word from the annotation vocabulary $w_i\in W^A$ to engage in a positive association $p(+|w_i)$ or being junk $p(J|w_i)$ by marginalizing over all annotators and object classes.
We then define the weighted relevance of the selected words $S$ as:
$Relevance(S) = \frac{\sum_{w_i \in S\cap W^A}p(+|w_i)}{\sum_{w_j \in W^A}p(+|w_j)}$, 
and similarly $Junk(S) = \frac{\sum_{w_i \in S\cap W^A}p(J|w_i)}{\sum_{w_j \in W^A}p(J|w_j)}$ for the junk score.
The final saliency score of $S$ is then defined as the average of both: $Saliency(S)=0.5(Relevance(S) + (1-Junk(S)))$.

\tblref{tab:words_perf} shows the performance of our approach and the baselines from the previous section.
While some of the baselines performed relatively well in getting a good attribute embedding, large portions of the selected words by these methods do not carry a clear semantic concept.
Our approach has a much higher relevance score while at the same time the lowest junk score among all baselines.
This indicates that the set of attributes discovered by our method correlates well with the human concept of semantic attributes.

\makeatletter{}
\begin{table}[t]
\centering
\scalebox{0.9}{
\begin{tabular}{r | c c | c c}

\toprule
\multirow{ 2}{*}{Model} & \multicolumn{2}{c|}{Attributes}		&	\multicolumn{2}{c}{Categories (DAP)} \\
						& 	Accuracy		&	AP				& 	Top1			&	AP \\
\midrule
\multicolumn{5}{l}{\textbf{Joint Model}} \\
w/o Linguistic Prior	&	55.2			&	22.4			&	30.4			&	19.0 	\\   w/~~ Linguistic Prior	&	\textbf{60.3}	&	\textbf{28.9}	&	\textbf{45.2}	&	\textbf{39.5} 	\\   \midrule
\multicolumn{5}{l}{\textbf{Attribute Model}} \\
w/o Association Opt.	&	74.8			&	64.1			&	51.4 			&	48.3	\\  w/~~	Association Opt.&	\textbf{76.9}	&	\textbf{68.2}	&	\textbf{55.9}	&	\textbf{54.2}	\\  
\bottomrule

\end{tabular}
}
\tablblvspace
\caption{Attribute prediction performance.}
\label{tab:attr_pred}
\tabvspace
\end{table}
 
\makeatletter{}
\begin{figure}[!t]
\centering
\begin{subfigure}[b]{0.47\linewidth}
    \centering
    \includegraphics[width=\linewidth]{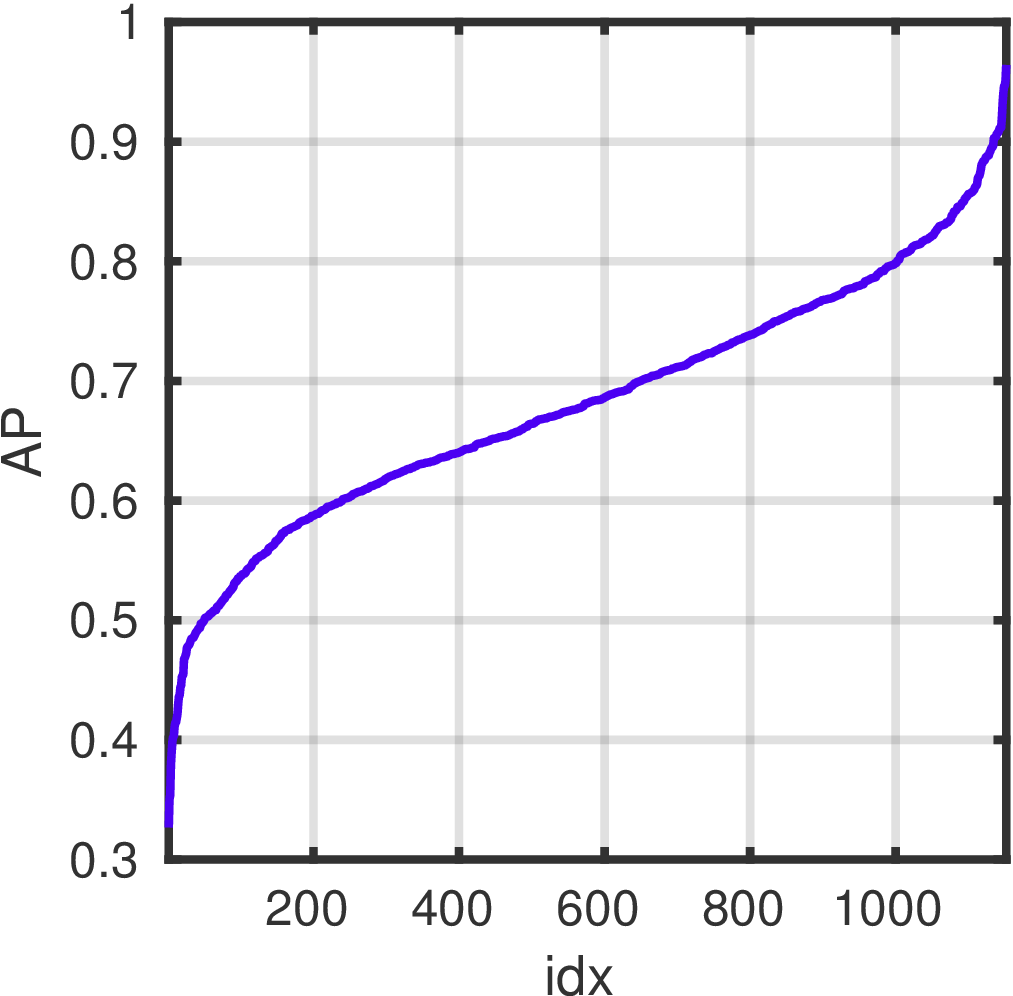}
    \caption{ }\label{fig:attr_ap}
\end{subfigure}\quad
\begin{subfigure}[b]{0.47\linewidth}
    \centering
    \includegraphics[width=\linewidth]{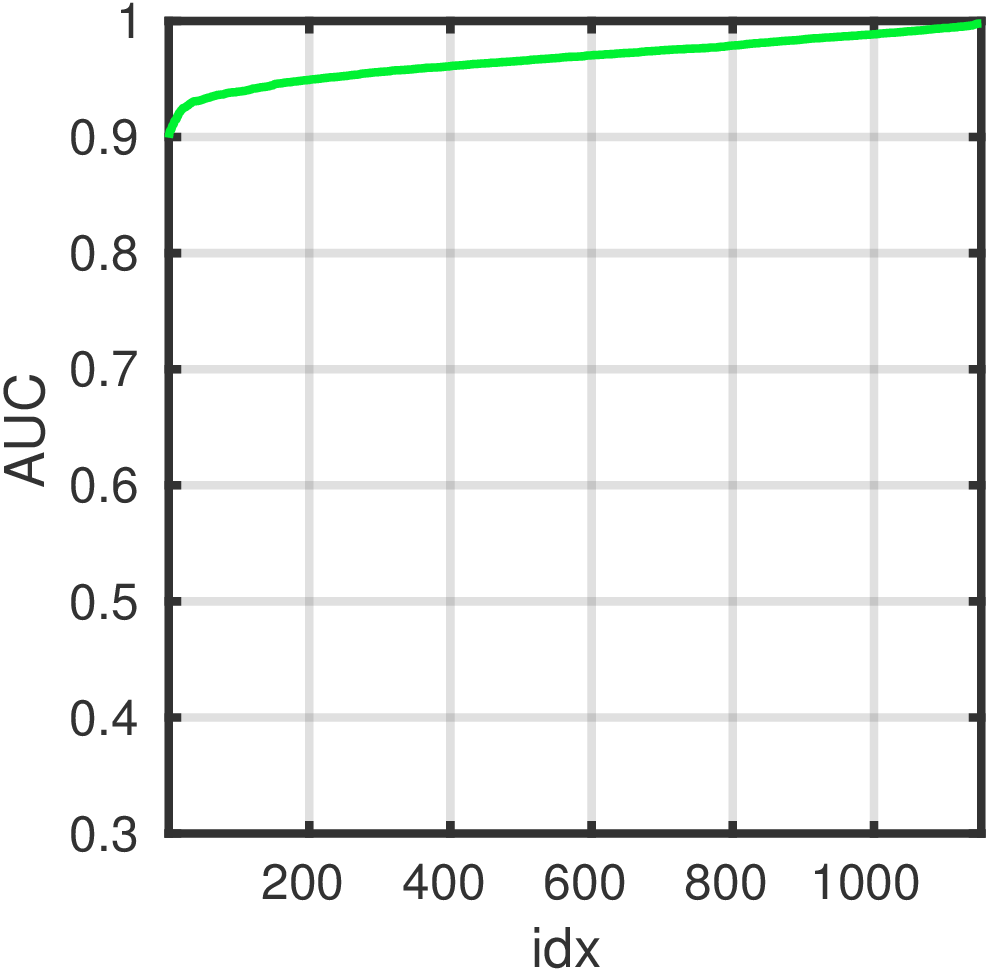}
    \caption{ }\label{fig:attr_auc}
\end{subfigure}
\figlblvspace
\caption{Performance of individual attributes in average precision (AP) and area under receiver operating characteristic (AUC).}
\label{fig:attr_apauc}
\figvspace
\end{figure} 
\subsection{Attribute prediction}
Having selected a set of salient attributes, we evaluate here the performance of our model in predicting these attributes in images.
\tblref{tab:attr_pred} shows the attribute prediction accuracy and average precision (AP).
It also reports the object Top1 classification accuracy and the AP based on the predicted attributes and when using the DAP model (\eqref{eq:dap}).

\paragraph{Joint Model}
In the first section of \tblref{tab:attr_pred}, it is interesting to see that regularizing the weights of the last fc layer with the language prior improves the performance of attribute predictions by 5\% in accuracy and 6\% in AP.
At the same time, it results in a boost in object classification Top1 accuracy by 15\%.
These results show that side information obtained from language has a significant impact on the performance of the deep model. 
Additionally, the unregularized network learns quite different associations between classes and attributes than those in $\mathbf{M}^l$.
Only 13\% of the positive associations in this case are shared with those learned from the textual description.
This indicates that the semantic link between the attributes and the classes is lost in this model.
In contrast, the regularized model preserves the semantics and retains more than 93\% of the positive associations in $\mathbf{M}^l$.
\paragraph{Attribute Model}
Finally, training the deep attribute model with the optimized associations $\mathbf{M}^*$ results in a better model compared to a one trained directly using $\mathbf{M}^l$.
This indicates that our joint model managed to account for some of the noise and missing data in $\mathbf{M}^l$.
The deep attribute model trained with $\mathbf{M}^*$ has higher attribute and object prediction performance.
Moreover, our deep attribute model achieves 75\% Top5 object classification accuracy, by predicting objects through the semantic attribute layer.
This is an impressive performance of the attribute model since it almost matches the performance of a deep model with the same architecture trained directly for object classification (80\% accuracy). 

\figref{fig:attr_apauc} shows the performance of the individual attributes.
Around 80\% of the attributes can be predicted with an average precision better than 0.6.

\makeatletter{}
\begin{table}[t]
\centering
\scalebox{0.8}{
\begin{tabular}{l c c c}

\toprule
Model 													&	Split	& 	200 labels	& 1000 labels \\
\midrule
Rohrbach \textit{et al.} \cite{Rohrbach2011}	&	A		& 	34.8 		& -  	\\ PST	\cite{Rohrbach2013}							&	A		& 	34.0		& -		\\
\ours 														&	A 		&	\textbf{46.1}		& \textbf{15.9}		\\ \ours - BT 												&	A 		&	\textbf{48.0}		& \textbf{20.2}		\\ \midrule
Mensink \textit{et al.} \cite{Mensink2012}		&	B		&	35.7 		& ~~1.9 \\
DeViSE \cite{Frome2013}							&	B		&	31.8 		& ~~9.0 \\
ConSE \cite{Norouzi2014}						&	B 		&	28.5		& -		\\ AMP (SR+SE) \cite{Fu2015}						&	B		&	41.0		& -		\\
\ours														& 	B 		&	\textbf{46.3}		& \textbf{15.2}		\\ \ours - BT													& 	B 		&	\textbf{49.0}		& \textbf{20.0}		\\ 
\midrule
\ours (w/o assoc. opt.)									&	C		&	45.8		& 	14.8	\\ \ours  													&	C 		& 	\textbf{48.1}		&	\textbf{16.9}	\\ 
\bottomrule

\end{tabular}
}
\tablblvspace
\caption{Zero-shot performance (Top5 accuracy) on 200 unseen classes from Imagenet.}
\label{tab:imagenet_zsl}
\tabvspace
\end{table}
 
\subsection{Zero-shot learning}
An important feature of semantic attributes is their ability to form a shared knowledge layer which can be transfered to unseen classes.
We evaluate here the performance of our discovered attributes in classifying unseen classes (\ie zero-shot learning).
While there is no standard zero-shot split in ImageNet, there are two common splits used in the literature and defined over the ILSVRC2010 classes, split A from \cite{Rohrbach2011} and B from \cite{Mensink2012}.
Both of them, split the classes into 800 seen and 200 unseen categories. 
We train our model as before while this time we use only the 800 seen classes of the respective split and we test on the remaining unseen classes.

\tblref{tab:imagenet_zsl} shows the Top5 accuracy of our model over the two splits (A \& B).
Our deep attribute model outperforms the state-of-the-art by 11\% on split A and by 5\% on split B.
Furthermore, we analyze the bias of our model toward seen classes similar to \cite{Frome2013}.
In this test setup, both the seen and unseen labels are considered as candidates when predicting the object category.
Our model achieves 15\% accuracy on split A \& B and shows much less bias compared to state of the art with 6\% improvement.
Additionally, if we assume the availability of test data as a batch (Ours-BT), we can get a better estimation of the mean and standard deviation for classifiers scores in \eqref{eq:dap}. 
This results in additional improvement of performance by 3\%.

Since in the zero-shot settings, we optimize the associations using only data from the seen categories, we analyze in the last section of \tblref{tab:imagenet_zsl} (split C) the performance of our model with and without association optimization. 
Here again, we find that the association optimization did not result in a biased performance towards the seen classes, rather it improved the model performance.
Overall, we see that optimizing the associations is beneficial in both within and across category prediction.

\makeatletter{}
\begin{table}[t]
\centering
\scalebox{0.75}{
\begin{tabular}{l l c c}

\toprule
Model 								&	Side Info.	& 	AwA 	& aPY \\
\midrule
\textbf{Supervised ZSL} & & &  \\
DAP~\cite{Lampert2013} (AlexNet)	&	A 					&	54.0		&	31.9	\\
DAP~\cite{Lampert2013} (GoogLeNet)	&	A 					&   59.5		&	37.1	\\
\midrule
\textbf{Unsupervised ZSL} & & & \\

DeViSE~\cite{Frome2013} 		& 	W					&	44.5		&	25.5	\\
Elhoseiny \etal ~\cite{Elhoseiny2013}& 	T					& 	55.3		&	30.2	\\
ConSE~\cite{Norouzi2014}		& 	W					&	46.1		&	22.0	\\			
SJE~\cite{Akata2015}			& 	G + H				&	60.1		&	-		\\
HAT~\cite{Al-Halah2015}			& 	H					&	59.7		&	31.1	\\
EZSL~\cite{Romera-Paredes2015}	&	T 					&	58.5		&	-		\\
Changpinyo \etal \cite{Changpinyo2016}			& 	W 					&   57.5		&   -		\\
Qiao \etal \cite{Qiao2016}		& 	T					&   66.5		&   -		\\
Xian \etal \cite{Xian2016}		& 	W + G + H			&   66.2		&   -		\\
CAAP~\cite{Al-Halah2016}		& 	W 					&   68.6		&   49.0	\\
\midrule
\ours (binary assoc.)			& 	T							  															&	77.3		& 	\textbf{57.6}	\\   \ours (continous assoc.)		&	T							  															&	\textbf{79.7}		& 	57.5	\\   \bottomrule

\end{tabular}
}
\tablblvspace
\caption{Zero-shot performance of various models on AwA and aPY. The supervised models use manually defined attributes (A), while the unsupervised approaches rely on other sources like word embeddings such as Word2Vec (W) \cite{Mikolov2013} and GloVe (G) \cite{Pennington2014}; hierarchy-based information (H) \cite{Miller1995} or textual description (T).}
\label{tab:ots_zsl}
\tabvspace
\end{table}
 
\paragraph{Across data sets zero-shot learning}
To compare the performance of our model that we learned in ImageNet with a manually selected attribute vocabulary, we evaluate our deep attribute model on two public data sets: 
1) Animals with Attributes (AwA) \cite{Lampert2009}: which has 50 animal classes split into 40 seen and 10 unseen categories with 84 predefined semantic attributes.
2) aPascal/aYahoo (aPY) \cite{Farhadi2009}: which has 32 classes split into 20 seen and 12 unseen with 64 semantic attributes.
We collect articles for each of the unseen categories to extract their associations to our discovered attribute vocabulary.
We consider both using the raw continuous associations (\ie tf$\cdot$idf values) and binary associations.
We test our model on the unseen categories on both data sets without any fine tuning of the trained deep model (off-the-shelf).

From \tblref{tab:ots_zsl} we see that our model outperforms all unsupervised zero-shot approaches. Compared to methods from \cite{Elhoseiny2013,Qiao2016} that used similar type of side information as ours, we have up to 13\% improvement. 
Moreover, our model outperforms a DAP model based on the manually defined attribute vocabulary and using image embeddings from an AlexNet model \cite{Krizhevsky2012} or even from GoogLeNet \cite{Szegedy2014}.
This demonstrates the impressive generalization properties of our model across data sets.

\makeatletter{}\begin{figure}[!t]
\centering
\begin{subfigure}[!b]{0.47\linewidth}
    \centering
    \includegraphics[width=\linewidth]{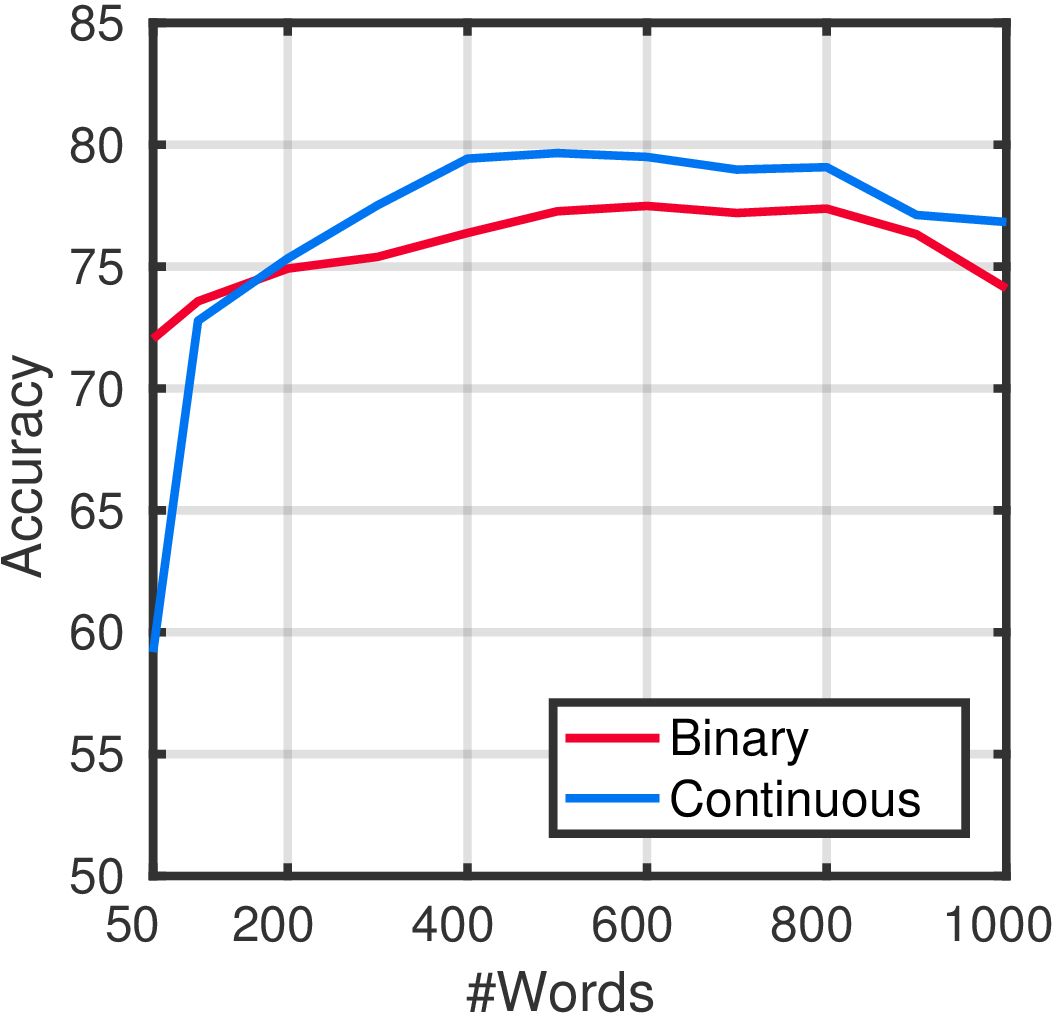}
    \caption{AwA}\end{subfigure}\quad
\begin{subfigure}[!b]{0.47\linewidth}
    \centering
    \includegraphics[width=\linewidth]{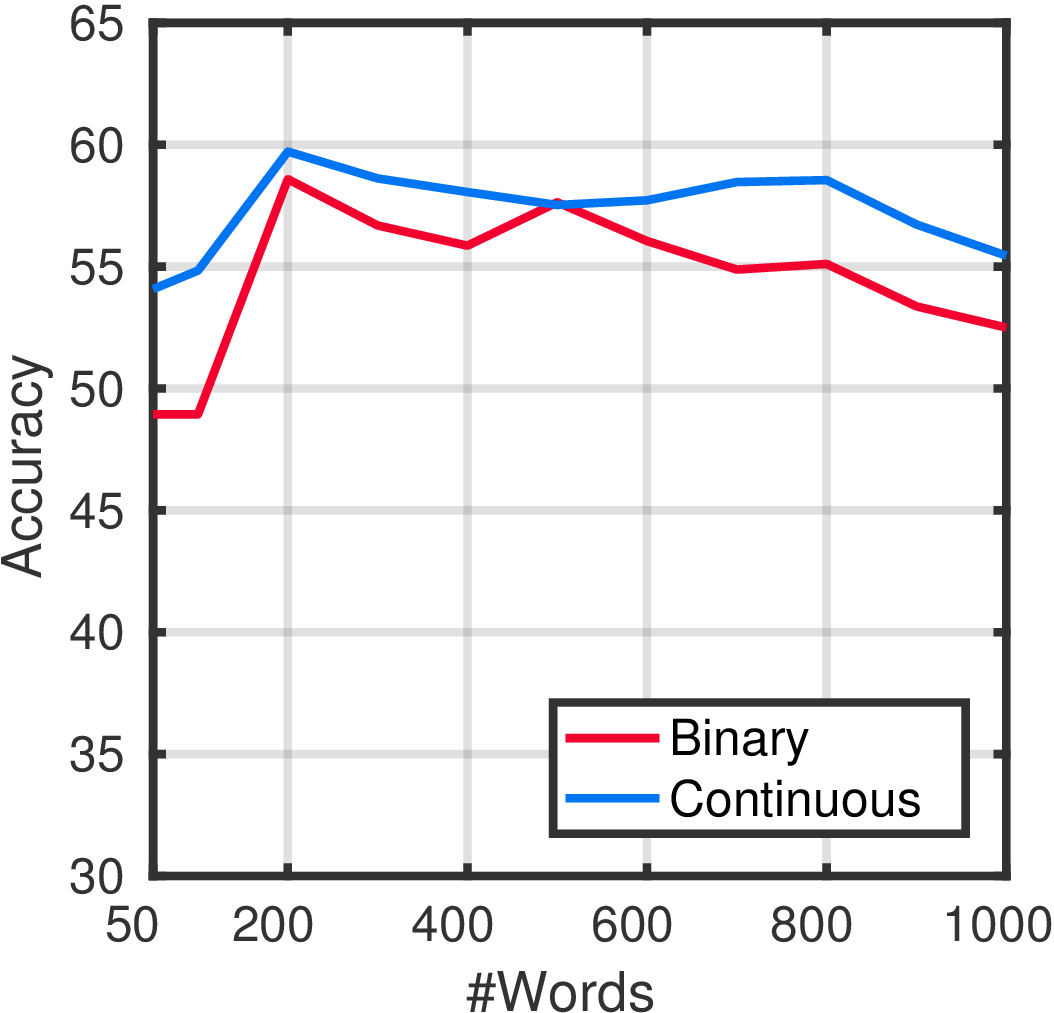}
    \caption{aPY}\end{subfigure}
\figlblvspace
\caption{Zero-shot performance with varying textual description lengths.}
\label{fig:text_length}
\figvspace
\end{figure} 
\paragraph{Text Length}
Here, we explore the effect of the article length on the prediction performance.
We vary the length of the considered section of the articles from 100 to 1000 words.
Then we extract the associations of the unseen classes in AwA and aPY from the truncated articles.

\figref{fig:text_length} shows the performance of the model in this case.
We notice that the optimal length of the article increases in correlation with the granularity of the categories in the data set.
For AwA which contains only animal classes, in average longer articles (400 to 600 words) are needed to sufficiently extract discriminant associations.
In contrast, categories in aPY are easier to separate with shorter articles (200 words).
Moreover, we see that most of the important attributes are mentioned quite early in the article, with performance degrading when we consider relatively long articles (more than 800 words).
In both data sets, we see that continuous associations outperform their binary counterpart in predicting the categories in most cases. 

\subsection{Discovered Attributes}
Using our model, we have discovered and learned 1636 semantic attributes describing 1360 categories with more than 1.2 million images from ImageNet (ILSVRC2010 \& ILSVRC2012).
This amounts to roughly 2 million class-attribute associations. In average, each attribute is shared between 29 categories, and each category has about 33 active attributes.
Some of the most shared attributes (with more than 100 categories) are \emph{water}, \emph{black}, \emph{red}, \emph{breed}, \emph{tail}, \emph{metal}, \emph{coat}, \emph{device}, \emph{hunt}, \emph{plastic}, \emph{yellow} and \emph{hair}.
Some of the least shared attributes (with less than 10 categories) are \emph{cassette}, \emph{cowboy}, \emph{pumpkin}, \emph{sweater}, \emph{convertible}, \emph{ballistic}, \emph{hump}, \emph{axe}, \emph{drilling}, \emph{laundry}, \emph{cash} and \emph{quilt}.

\section{Conclusion}
\makeatletter{}We propose a novel end-to-end approach to discover and learn attributes at a large scale form textual descriptions.
Our model discovers a salient, diverse and discriminative set of attribute vocabulary that correlates well with human understanding of semantic attributes.
Moreover, in order to account for noise and missing data in the text corpora, we propose to use a linguistic prior in a joint deep model to optimize the class-attribute associations.
In an evaluation on ImageNet, we show that our deep attribute model is able to learn and predict semantic attributes with high accuracy for a thousand categories.
Our model outperforms the state-of-the-art in unsupervised zero-shot learning and it generalizes well across data sets.

\balance
{\reffontsize
\bibliographystyle{ieee}
\bibliography{cvpr17}
}
\end{document}